\documentclass[sigconf]{acmart}

\usepackage{multirow}
\usepackage{colortbl}
\usepackage{subcaption}
\usepackage{siunitx}
\usepackage{amsmath}
\usepackage{enumitem}
\usepackage{balance}

\AtBeginDocument{%
  }

\copyrightyear{2025}
\acmYear{2025}
\setcopyright{acmlicensed}\acmConference[FPGA '25]{Proceedings of the 2025 ACM/SIGDA International Symposium on Field Programmable Gate Arrays}{February 27-March 1, 2025}{Monterey, CA, USA}
\acmBooktitle{Proceedings of the 2025 ACM/SIGDA International Symposium on Field Programmable Gate Arrays (FPGA '25), February 27-March 1, 2025, Monterey, CA, USA}
\acmDOI{10.1145/3706628.3708877}
\acmISBN{979-8-4007-1396-5/25/02}

\author{Alireza Khataei}
\affiliation{%
  \institution{Department of Electrical and Computer Engineering\\University of Minnesota}
  \city{Minneapolis}
    \state{MN}
  \country{USA}}
\email{khata014@umn.edu}

\author{Kia Bazargan}
\affiliation{%
  \institution{Department of Electrical and Computer Engineering\\University of Minnesota}
  \city{Minneapolis}
  \state{MN}
  \country{USA}}
\email{kia@umn.edu}

\begin{document}

\title[\parbox{0.477\textwidth}{TreeLUT: An Efficient Alternative to Deep Neural Networks for Inference Acceleration Using Gradient Boosted Decision Trees}]{TreeLUT: An Efficient Alternative to Deep Neural Networks for Inference Acceleration Using Gradient Boosted Decision Trees}


\begin{abstract}
Accelerating machine learning inference has been an active research area in recent years. In this context, field-programmable gate arrays (FPGAs) have demonstrated compelling performance by providing massive parallelism in deep neural networks (DNNs). Neural networks (NNs) are computationally intensive during inference, as they require massive amounts of multiplication and addition, which makes their implementations costly. Numerous studies have recently addressed this challenge to some extent using a combination of sparsity induction, quantization, and transformation of neurons or sub-networks into lookup tables (LUTs) on FPGAs. Gradient boosted decision trees (GBDTs) are a high-accuracy alternative to DNNs in a wide range of regression and classification tasks, particularly for tabular datasets. The basic building block of GBDTs is a decision tree, which resembles the structure of binary decision diagrams. FPGA design flows are heavily optimized to implement such a structure efficiently. In addition to decision trees, GBDTs perform simple operations during inference, including comparison and addition. We present TreeLUT as an open-source tool for implementing GBDTs using an efficient quantization scheme, hardware architecture, and pipelining strategy. It primarily utilizes LUTs with no BRAMs or DSPs on FPGAs, resulting in high efficiency. We show the effectiveness of TreeLUT using multiple classification datasets, commonly used to evaluate ultra-low area and latency architectures. Using these benchmarks, we compare our implementation results with existing DNN and GBDT methods, such as DWN, PolyLUT-Add, NeuraLUT, LogicNets, FINN, hls4ml, and others. Our results show that TreeLUT significantly improves hardware utilization, latency, and throughput at competitive accuracy compared to previous works. For instance, it achieves an accuracy of around 97\% on the MNIST dataset while delivering around 4 to 101 times lower hardware cost in terms of area-delay product than recent LUT-based NNs.
\end{abstract}



\maketitle

\section{Introduction}
Deep neural networks (DNNs) are powerful machine learning models deployed in a wide range of applications such as image recognition, natural language processing, and recommender systems. However, the computational and memory demands of DNN models complicate their deployment in ultra-low latency and high throughput tasks, especially on resource-constraint edge devices. Field-programmable gate arrays (FPGAs) are a promising solution to deploy DNNs in such applications by providing massive parallelism and customizable architectures to accelerate them during inference.

Neural networks (NNs) heavily perform multiply-accumulate operations and nonlinear activation functions. Moreover, they typically need to access large amounts of data, such as weights and biases. These complexities impose constraints on their hardware implementations and come at the expense of high hardware costs. Numerous techniques have been proposed in the literature to address NN implementation challenges through binarization, quantization, and transformation of neurons or sub-networks into lookup tables (LUTs). Recently, several LUT-based NN methods have been proposed as a promising solution to compress NN models for FPGA implementations. Examples of such methods include PolyLUT-Add~\cite{10705569}, NeuraLUT~\cite{10705559}, PolyLUT~\cite{10416099}, LogicNets~\cite{9221584}, LUTNet~\cite{8735521}, and NullaNet~\cite{nazemi2018nullanet}. The intuition behind these methods is that a neuron or a sub-network can be implemented as a black box mapped to LUTs. For example, LogicNets maps each neuron of a neural network to one LUT. In other words, it fully tabulates the output values of a neuron for all possible input values. As a result, the multiplication, addition, and activation function of a neuron are all absorbed in a LUT that directly maps its inputs to its output. These methods typically need to heavily quantize values going from LUTs to LUTs and exploit the sparsity of networks to fit neurons or sub-networks into small LUTs, which might hurt the accuracy of the baseline NN models while not fully leveraging the unique strengths of FPGA synthesis tools. In addition, training such models, using sparsity patterns, and mapping them to LUTs are relatively time-consuming.

Recent LUT-based NN works motivated us to explore other machine learning models to find a baseline that could easily leverage the unique features of FPGAs with minimal transformations. Gradient boosted decision trees (GBDTs) are an efficient alternative to DNNs on FPGAs, mainly because of the following two reasons.
\begin{itemize}[left=0pt]
    \item \textbf{Quality}: GBDTs are powerful machine learning models, used in a wide range of regression and classification tasks. GBDTs have demonstrated competitive accuracy compared to NNs in various tasks, particularly on tabular data~\cite{NEURIPS2023_f06d5ebd}. For example, XGBoost~\cite{10.1145/2939672.2939785} -- a scalable implementation of the GBDT algorithm -- has been frequently used in machine learning challenges. At a Kaggle competition, XGBoost models were deployed in the majority of the winning solutions, whereas DNNs were the second-most popular method~\cite{10.1145/2939672.2939785}.
    
    \item \textbf{Perfect Fit for FPGAs}: GBDTs mostly consists of decision trees, which resemble binary decision diagrams (BDDs). These BDD-like structures nicely map to LUTs, as FPGA design flows are highly optimized to implement such structures efficiently. Moreover, these models perform simple operations during inference, which makes their hardware implementations more efficient.
\end{itemize}  


Quantization and implementation of GBDTs on FPGAs have been studied in the literature. For example, \cite{electronics10030314} and \cite{9786085} have proposed GBDT inference acceleration methods. However, they use complex memory-based architectures. On the other hand, hls4ml (Summers et al.)~\cite{Summers_2020} has proposed a tool, called Conifer, to convert GBDT models into either VHDL RTL code or C++ HLS code using a LUT-based architecture. Although their HLS design benefits from more efficient pipelining than their RTL design, HLS synthesis time increases significantly as the complexity of the baseline GBDT model increases~\cite{Summers_2020}. This tool also applies a post-training quantization scheme using a fixed-point representation to lower hardware costs. Alsharari et al.\cite{10652593} have recently shown that Conifer has limited precision and proposed a quantization-aware training scheme to quantize GBDTs for integer-only inference. However, their scheme introduces significant complexities into the training process, and their method still relies on Conifer and HLS tools for hardware implementations.

In this paper, we present TreeLUT\footnote{The TreeLUT tool is available at \textcolor{blue}{\textbf{\url{https://github.com/kiabuzz/TreeLUT}}} (DOI:~10.5281/zenodo.14497551).} as an open-source tool for implementing GBDTs in hardware. To be more specific, TreeLUT is a Python library to quantize a GBDT model and directly convert it into Verilog hardware files without relying on HLS tools. It primarily utilizes LUTs with no BRAMs or DSPs on FPGAs, resulting in low area and low latency. TreeLUT uses an efficient quantization scheme, hardware architecture, and pipelining strategy. For the quantization, it uses a combination of pre-training and post-training quantization approaches without using complex quantization-aware training techniques, which makes the training efficient and straightforward. Furthermore, it tries to quantize decision trees using fewer number of bits than a target bitwidth by reforming inference equations, which makes hardware implementations less costly. For the hardware architecture, it uses a fully unrolled 3-layer architecture, making it modular, scalable, and efficient in terms of hardware costs. Finally, it inserts pipelining stages in the architecture to achieve lower latency and higher throughput, without hindering synthesis flow from efficient optimizations of the decision trees as combinational circuits.

We evaluated TreeLUT on FPGAs on the MNIST, JSC, and NID classification datasets, that are used for the evaluation of ultra-low latency and area architectures~\cite{10416099}. These datasets have been widely used for architectural evaluations by recent LUT-based NNs in terms of accuracy and hardware costs. The post-place\&route implementation results on FPGAs demonstrate that TreeLUT has outstanding performance on these datasets at competitive accuracy compared to previous works~\cite{pmlr-v235-bacellar24a, 8596871, 10705569, 10705559, 10416099, 10.1145/3020078.3021744, Ngadiuba_2021, fahim2021hls4ml, Summers_2020, 10652593, MUROVIC20211, 9221584}.

The rest of the paper is structured as follows. In Section~\ref{sec:methodology}, the background on decision trees and gradient boosting is provided. Moreover, the quantization scheme, hardware architecture and pipelining strategy of TreeLUT are presented. In Section~\ref{sec:toolflow}, an overview of the TreeLUT tool flow is outlined. Finally, the experimental setup and evaluation results are provided and discussed in Section~\ref{sec:evaluation}.

\section{Methodology}
\label{sec:methodology}

\subsection{Background}

\subsubsection{\textbf{Decision Trees}} 
\begin{figure} [] 
	\centering
		\includegraphics[scale=.675] {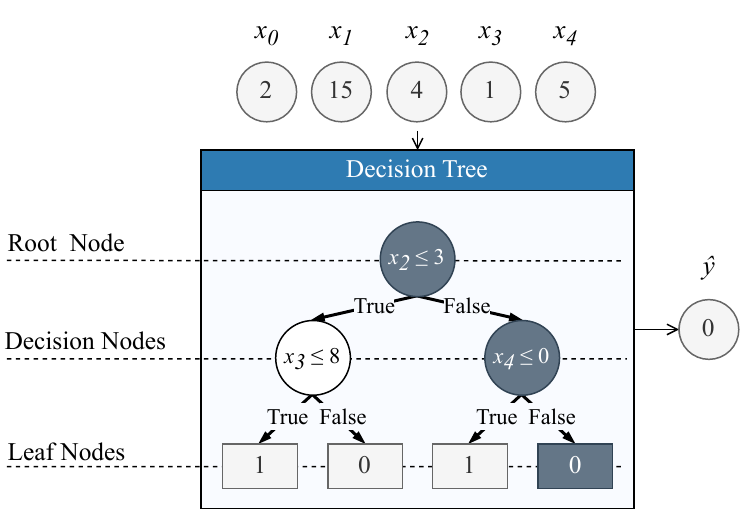} 
	\caption{An example of a single decision tree during inference for a binary classification task. The leaves are discrete values, directly representing predicted class labels. There could be multiple levels of decision nodes.}
    \vspace{-8pt}
 \label{fig:decisiontree}
\end{figure}

Decision trees are machine learning models, used in different tasks, including regression and classification. They have a binary tree structure, consisting of internal nodes and leaf nodes. During inference, specific input features are compared against fixed thresholds at the internal nodes, and different paths are taken based on the outcomes of the comparisons, which finally lead to a single leaf node. In classification tasks, the leaves are typically discrete values, representing predicted class labels. In regression tasks, however, the leaves are continuous values. During training, the decision tree is structured, the features and their corresponding thresholds are determined, and the leaf values are assigned. Fig.~\ref{fig:decisiontree} shows a trained decision tree during inference for a binary classification task. As seen, the leaves are discrete values, representing the predicted class labels. In this example, the arbitrary input feature vector $X = [2, 15, 4, 1, 5]$ is fed to the decision tree, which is eventually classified as Class 0.

\subsubsection{\textbf{Gradient Boosting}}
\label{sec:gbdt}
Using a single decision tree is inefficient, demonstrating poor accuracy in complex tasks. Gradient boosting is an ensemble method that trains a series of ``weak'' predictors in an additive form to build a more accurate model~\cite{4a848dd1-54e3-3c3c-83c3-04977ded2e71}. GBDT is a gradient-boosting variation that uses decision trees as its weak predictors. It achieves state-of-the-art performance in various regression and classification tasks~\cite{NIPS2017_6449f44a}. 

There are various implementations of GBDT in the literature, including CatBoost~\cite{NEURIPS2018_14491b75}, LightGBM~\cite{4a848dd1-54e3-3c3c-83c3-04977ded2e71}, and XGBoost~\cite{10.1145/2939672.2939785}, each of which provides different benefits in terms of efficiency, scalability, and others. As these variations fundamentally follow a similar mathematical procedure during inference, our method can be applied to any of them. However, we designed our TreeLUT tool based on XGBoost, as it is widely used as a high-performance and scalable tool for implementing GBDT models. Throughout the paper, GBDT refers to XGBoost, regardless of the similarities and differences among the various GBDT implementations.

During the training of a GBDT model,  a set of decision trees is trained iteratively, each of which is fitted to the residuals of the previous ones. These decision trees have continuous values on leaves for both regression and classification tasks, as they predict continuous scores during inference, which are accumulated to form the final prediction value as follows.
 \begin{equation}
F(X) = f_0 + \sum_{m = 1}^{M} {f}_{m}(X)
\label{eq:xgb}
\end{equation}
where the hyperparameter $M$ is the number of decision trees in the ensemble model. In addition, the constant $f_0$ is an initial prediction score, and $f_m(X)$ is the prediction score returned by the $m$-th decision tree given input $X$. In classification tasks, the summation can be turned into probability values using activation functions. 

\begin{figure} [] 
	\centering
		\includegraphics[scale=.56] {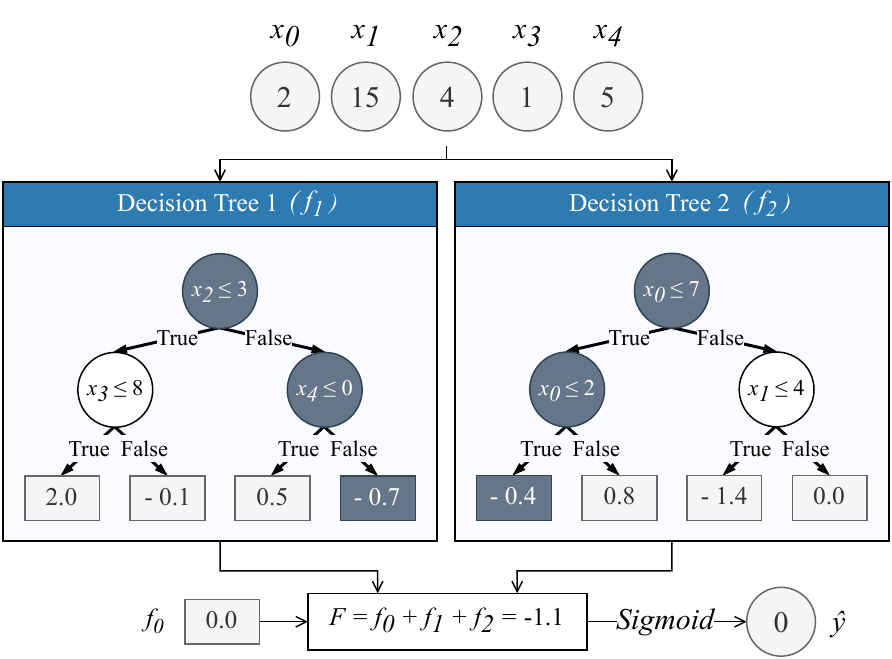} 
	\caption{An example of a GBDT model during inference for a binary classification task. The leaves are continuous values, representing predicted scores. The final class prediction is performed based on the summation of the scores. There could be more decision trees with multiple levels of decision nodes.}
    \vspace{-8pt}
 \label{fig:xgb}
\end{figure}

In a binary classification task, $F(x)$ is turned into a probability using a \textit{sigmoid} function and then used for class predictions. Fig.~\ref{fig:xgb} shows a trained GBDT model during inference for a binary classification task. As seen, the leaves are continuous values, representing the prediction scores. In this example, the arbitrary input feature vector $X = [2, 15, 4, 1,5]$ is fed to the decision trees, and they independently predict the scores $f1 = -0.7$ and $f2 = -0.4$. Then, these values and the initial prediction score $f_0 = 0.0$ are added together to form the final prediction value $F = -1.1$. As the final score turns out to be a negative value in this example, it causes the output of the \textit{sigmoid} function to be less than 0.5, translated to Class 0 as the class prediction. In multiclass classification, however, a one-vs-all strategy is usually used. As a result, a separate set of $M$ decision trees are trained for each class, which are added across the classes and then passed to a \textit{softmax} function for class predictions.

\subsection{Quantization}
Training a GBDT model results in independent decision trees with fixed internal decision node thresholds and leaf values. In this section, we discuss how to quantize these values for efficient hardware implementations while preserving high accuracy.

\subsubsection{\textbf{Threshold Quantization}}
In each decision tree during inference, a subset of input features are compared against fixed internal decision thresholds. If a GBDT model is trained on a dataset with floating-point feature values, the training algorithm obtains the thresholds in floating-point as well, and quantizing them might introduce significant approximation errors. For this reason, unlike previous research efforts, we simply quantize the feature values before training and then run the learning algorithm on the quantized data. This approach helps the boosting algorithm to find optimal quantized thresholds by itself without the need for conventional quantization-aware training or post-training quantization techniques. In other words, the boosting algorithm chooses the most informative features and their corresponding thresholds, given the information loss introduced by the quantization of the input features. As a result, we use a uniform quantization scheme to quantize data as a pre-processing step.  First, a min-max normalization is applied to input features to linearly scale them to the range [0, 1].
 \begin{equation*}
  \begin{aligned}
X_{normalized} &= \frac{X - min(X)}{max(X) - min(X)}
  \end{aligned}
\end{equation*}
Then, the normalized features are quantized into $w\_feature$ bits, where $w\_feature$ is a quantization hyperparameter.
 \begin{equation*}
  \begin{aligned}
X_{quantized} &= round(X_{normalized}\times (2^{w\_feature}-1))
  \end{aligned}
\end{equation*}

\subsubsection{\textbf{Binary Classification Leaf Quantization}}
\label{sec:leaf}
In binary classification, a GBDT model consists of a total number of $M$ decision trees, where $M$ is a boosting hyperparameter. During inference, an input feature vector $X$ is fed into the decision trees, each of which independently returns a prediction score in floating-point. The final class prediction is performed as follows.
\begin{equation}
\hat{y} = \left\{
         	 \begin{array}{lr}
            		1 &  F(X) \geq 0\\
            		0 &  F(X) < 0
         	 \end{array}
        \right.;~F(X) = f_0 + \sum_{m = 1}^{M} {f}_{m}(X)
\label{eq:binary}
\end{equation}
where $F(X)$ denotes the final prediction value, $f_0$ is an initial prediction score, and ${f}_m(X)$ denotes the prediction score, i.e. the leaf value, returned by the $m$-th decision tree given the input $X$. For the leaf quantization, we subtract the minimum leaf value of each decision tree from all their leaves and rewrite the equation of $F(X)$ as follows. 
\begin{equation*}
    \forall m \in\{1,\cdots M\}, \quad minLeaf_m = \min_{X}({f}_{m}(X))
\end{equation*}
where $\min_{{X}}({f}_{m}({X}))$ is the minimum leaf value in the decision tree~${f}_{m}$.
{\small \begin{equation}
 \begin{aligned}
F(X) &= f_0 + \sum_{m = 1}^{M} {f}_{m}(X)\\
    &= f_0 + \sum_{m = 1}^{M} [{f}_{m}(X) - minLeaf_m] +  \sum_{m = 1}^{M} minLeaf_m\\
    &= [f_0 + \sum_{m = 1}^{M} minLeaf_m] + \sum_{m = 1}^{M} [{f}_{m}(X) - minLeaf_m] \\
F(X) &= b + \sum_{m = 1}^{M} {f}^{\prime}_{m}(X)
\end{aligned}
\label{eq:binary_min}
\end{equation}}
In this equation, ${f}^{\prime}_{m}(X)$ returns a non-negative value as opposed to ${f}_{m}(X)$, as we shifted the range of each decision tree into a positive range. The constant $b$, called a ``bias'', is often a negative value, as it is the summation of the initial prediction score and the minimum values of the leaves.\footnote{Table~\ref{tbl:numExample} (second row) shows the changes in the leaf values of Fig.~\ref{fig:xgb}.} If it turns out to be positive, it means that the classifier classifies any inputs as Class 1, according to Eq.~\ref{eq:binary}. In the next step, we scale $F(x)$ by a positive scaling factor as follows.\footnote{Table~\ref{tbl:numExample} (third row) shows the changes after scaling by $binaryScale=7/2.7 = 2.59$. Here, $w\_tree=3$ and $\max_{i, X}({f}^{\prime}_{i}(X)) = 2.7$.}
\begin{equation*}
    binaryScale = \frac{2^{w\_tree}-1}{\max_{i, X}({f}^{\prime}_{i}(X))}
\end{equation*}
where $w\_tree$ is a quantization hyperparameter, and $\max_{i, X}({f}^{\prime}_{i}(X))$ is the maximum leaf value across all the decision trees $f^\prime$.
\begin{equation}
 \begin{aligned}
F^{\prime}(X)   &= F(X)\times binaryScale\\
                &= b\times binaryScale \; + \sum_{m = 1}^{M} {f}^{\prime}_{m}\times binaryScale\\
F^{\prime}(X)   &= b^{\prime} + \sum_{m = 1}^{M} {f}^{\prime\prime}_{m}(X)
\end{aligned}
\label{eq:binary_scale}
\end{equation}
As the scaling factor is positive, we can use $F^{\prime}(X)$ instead of $F(X)$ in Eq.~\ref{eq:binary} without introducing any approximation errors. 

\begin{equation}
\hat{y} = \left\{
         	 \begin{array}{lr}
            		1 &  F(X) \geq 0\\
            		0 &  F(X) < 0
         	 \end{array}
        \right. = \left\{
         	 \begin{array}{lr}
            		1 &  F^{\prime}(X) \geq 0\\
            		0 &  F^{\prime}(X) < 0
         	 \end{array}
        \right.
\label{eq:binary_new}
\end{equation}
For the quantization of the bias and leaf values, we use the rounding function to approximately write $F^{\prime}(X)$ as follows.
\begin{equation}
 \begin{aligned}
&F^{\prime}(X) \approx QF(X) = round(b^{\prime}) + \sum_{m = 1}^{M} round({f}^{\prime\prime}_{m}(X))\\[-8pt]
& QF(X) = qb + \sum_{m = 1}^{M} {qf}_{m}(X)
\end{aligned}
\label{eq:quantize}
\end{equation}
where ${qf}_{m}$ is the quantized decision tree, whose leaf values are quantized into $w\_tree$ bits, and $qb$ is the quantized bias, which is still a negative value and typically fits in more than $w\_tree$ bits.\footnote{Table~\ref{tbl:numExample} (last row) shows the changes after rounding.} As seen, we use the local minimum values and the global maximum value for shifting and scaling the leaves, respectively. This approach ensures that the minimum leaf value in each quantized decision tree is 0. In addition, the maximum leaf values in many of the quantized decision trees fit in fewer than $w\_tree$ bits, as we apply the global maximum value for the scaling.\footnote{In practice, a large number of decision trees might use one or two fewer bits than $w\_tree$, because their leaf values use only half or a quarter of the maximum range of all the leaves from all decision trees.} Had we used the global minimum value for shifting, that would have created offsets in each quantized decision tree, which could have increased their bitwidths and made the hardware implementations more costly. Putting everything together, we can use the following equation instead of Eq.~\ref{eq:binary} to approximately predict the classes in binary classification.
\begin{equation}
\hat{y} \approx \left\{
         	 \begin{array}{lr}
            		1 &  QF(X) \geq 0\\
            		0 &  QF(X) < 0
         	 \end{array}
        \right.;~QF(X) = qb + \sum_{m = 1}^{M} {qf}_{m}(X)
        \label{eq:binary_final}
\end{equation}
In Eq.~\ref{eq:binary}, \ref{eq:binary_new}, and \ref{eq:binary_final}, the threshold $0$ equals to the typical classification threshold $0.5$ in the \textit{sigmoid} function, and it can be adjusted in case of class imbalance. Any adjusted thresholds can be combined with the bias $b$ and quantized as a single $qb$. The XGBoost library, however, supports a hyperparameter, which automatically adjusts the prediction scores to address the class imbalance. Table~\ref{tbl:numExample} shows the leaf quantization for the example GBDT model in Fig.~\ref{fig:xgb}. The leaf values are quantized into 3 bits in this example.

\begin{table}[H]

\caption{Numeric example of equations \ref{eq:binary_min} - \ref{eq:quantize}.}
\vspace{-5pt}
\renewcommand{\arraystretch}{1.45}
\resizebox{\linewidth}{!}{ 
\begin{tabular}{|l|r|rrrr|rrrr|}
\hline
\multicolumn{1}{|c|}{\textbf{Step}} & \textbf{Bias} & \multicolumn{4}{c|}{\textbf{Decision Tree 1}} & \multicolumn{4}{c|}{\textbf{Decision Tree 2}} \\ \hline \hline
\rowcolor[HTML]{EFEFEF} 
Original & 0.00 & 2.00 & -0.10 & 0.50 & -0.70 & -0.40 & 0.80 & -1.40 & 0.00 \\ 
After Eq. \ref{eq:binary_min} & -2.10 & 2.70 & 0.60 & 1.20 & 0.00 & 1.00 & 2.20 & 0.00 & 1.40 \\ 
\rowcolor[HTML]{EFEFEF} 
After Eq. \ref{eq:binary_scale} & -5.44 & 7.00 & 1.56 & 3.11 & 0.00 & 2.59 & 5.70 & 0.00 & 3.63 \\
After Eq. \ref{eq:quantize} & -5 & 7 & 2 & 3 & 0 & 3 & 6 & 0 & 4 \\ \hline
\end{tabular}
}
\label{tbl:numExample}
\end{table}


\subsubsection{\textbf{Multiclass Classification Leaf Quantization}}
\label{sec:leafMultiClass}
In multiclass classification, a GBDT model consists of an $N\times M$ array of decision trees, where $N$ is the number of classes, and $M$ is the number of decision trees per class. During inference, an input feature vector $X$ is fed into the decision trees, each of which independently returns a prediction score in floating-point. The final class prediction is performed as follows.
 \begin{equation}
\hat{y} = argmax_n(F_n(X));~F_n(X) = f_0 + \sum_{m = 1}^{M} {f}_{n, m}(X)
\label{eq:multi}
\end{equation}
where $F_n(X)$ denotes the final prediction value of the $n$-th class, $f_0$ is an initial prediction score, and ${f}_{n, m}(X)$ denotes the leaf value returned by the $m$-th decision tree in the $n$-th class given input $X$. As seen, there are $N$ separate final prediction values in multiclass classification. Using a similar approach, we quantize each $F_n(X)$ by shifting the leaves with respect to their local minimum values and scale them by a single factor using the global maximum value across the decision trees in all the classes, which results in a separate bias for each class. As a result, we rewrite $F_n(X)$ as follows.

\begin{equation*}
    \forall m \in\{1,\cdots M\}, n \in\{1,\cdots N\},\quad minLeaf_{n,m} = \min_{X}({f}_{n,m}(X))
\end{equation*}
where $\min_{X}({f}_{n, m}(X))$ is the minimum leaf value in the decision tree ${f}_{n, m}$. 
{\small \begin{equation}
 \begin{aligned}
F_n(X) &= f_0 + \sum_{m = 1}^{M} {f}_{n, m}(X)\\
    &= f_0 + \sum_{m = 1}^{M} [{f}_{n, m}(X) - minLeaf_{n,m}] +  \sum_{m = 1}^{M} minLeaf_{n,m}\\
    &= [f_0 + \sum_{m = 1}^{M} minLeaf_{n,m}] + \sum_{m = 1}^{M} [{f}_{n, m}(X) - minLeaf_{n,m}] \\
F_n(X) &= b_n + \sum_{m = 1}^{M} {f}^{\prime}_{n, m}(X)
\end{aligned}
\label{eq:multi_min}
\end{equation}}
In the equation above, the constant $b_n$ is typically a negative value. However, as we use the \textit{argmax} function in multiclass classification, we can add a constant value to all the biases to make them positive. In the next step, we scale all the $F_n(X)$ by a single positive factor as follows.
\begin{equation*}
    multiScale = \frac{2^{w\_tree}-1}{\max_{i, j, X}({f}^{\prime}_{i, j}(X))}
\end{equation*}
where $\max_{i, j, X}({f}^{\prime}_{i, j}(X))$ is the maximum value across all the decision trees $f'$ in all the classes.
\begin{equation}
 \begin{aligned}
F'_n (X) &= F_n(X)\times multiScale\\ &= b_n\times multiScale + \sum_{m = 1}^{M} {f}_{n, m}\times multiScale\\
F'_n (X) &= b^{\prime}_n + \sum_{m = 1}^{M} {f}^{\prime\prime}_{n, m}(X)
\end{aligned}
\label{eq:multi_scale}
\end{equation}
As the scaling factor is positive and constant across the classes, we can use $F^{\prime}(X)$ instead of $F(X)$ in Eq.~\ref{eq:binary} without introducing any approximation errors.
\begin{equation*}
\hat{y} = argmax_n(F_n(X)) = argmax_n(F'_n(X))
\end{equation*}
Next, we use the rounding function to quantize the bias and leaves as follows.
\begin{equation*}
 \begin{aligned}
&F'_n(X) \approx QF_n(X) = round(b^{\prime}_n) + \sum_{m = 1}^{M} round({f}^{\prime\prime}_{n, m}(X))\\[-8pt]
& QF_n(X) = qb_n + \sum_{m = 1}^{M} {qf}_{n, m}(X)
\end{aligned}
\end{equation*}
Finally, we can use the following equation instead of Eq.~\ref{eq:multi} to approximately predict the classes in the multiclass classification task.
\begin{equation}
\hat{y} \approx argmax_n(QF_n(X));~QF_n(X) = qb_n + \sum_{m = 1}^{M} {qf}_{n, m}(X)
\label{eq:multi_final}
\end{equation}

\subsection{Hardware Architecture}
\begin{figure} [] 
	\centering
	\includegraphics[scale=.57] {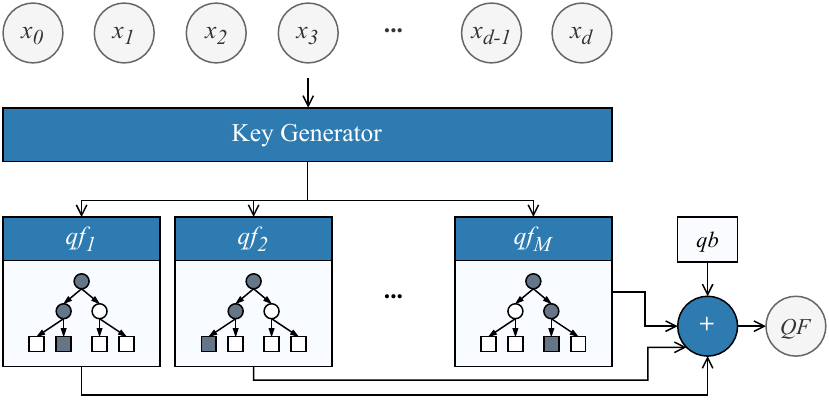} 
	\caption{The overall TreeLUT architecture for binary classification tasks.}
 \label{fig:binary}
\end{figure}

\begin{figure} [] 
	\centering
		\includegraphics[scale=.57] {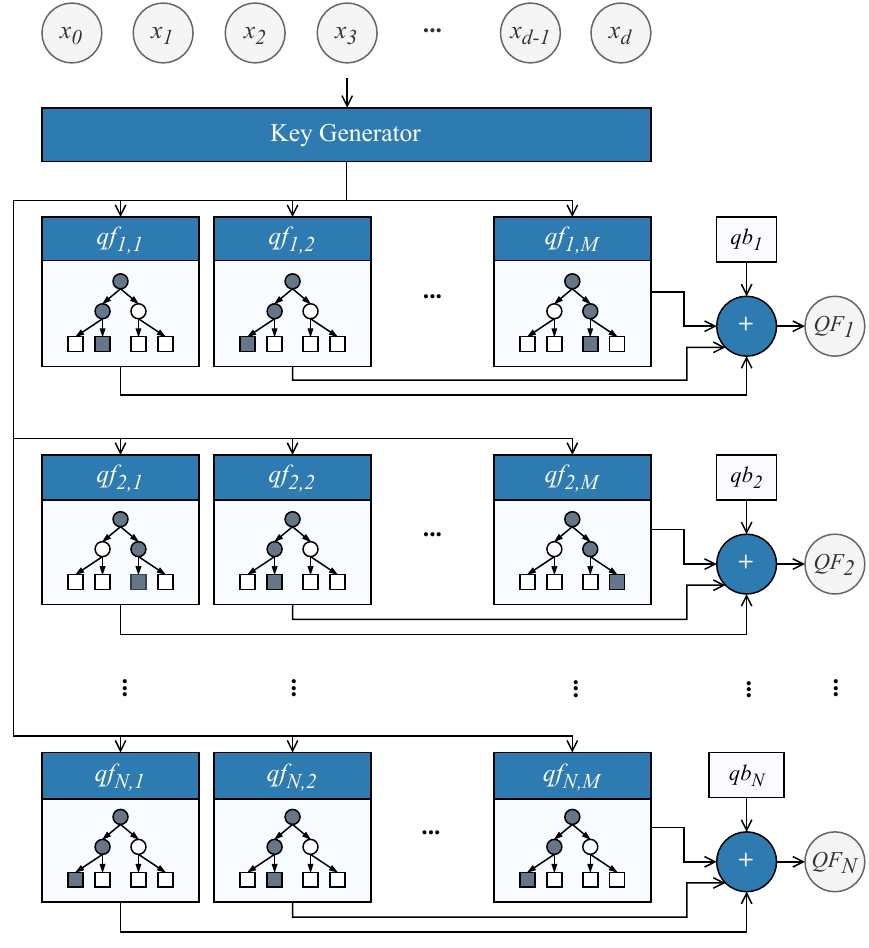} 
	\caption{The overall TreeLUT architecture for multiclass classification tasks.}

 \label{fig:multi}
\end{figure}

TreeLUT consists of three layers: key generator, decision trees, and adder trees. At the first level, decision nodes are compared to threshold values. At the second level, quantized decision trees are implemented. In the last layer, the values of quantized decision trees are added together using adder trees. Fig.~\ref{fig:binary} and Fig.~\ref{fig:multi} show the overall architecture of TreeLUT for binary and multiclass classification, respectively. Hardware implementation of tree-based models is not a complex task. Regardless of quantization schemes and particular types of tree-based models, there are similarities between our architecture and previous works on the hardware implementations of such tree-based models~\cite{Summers_2020, 9933867, 10653667, 7471178, 10.1007/978-3-319-03889-6_14, 10.1007/978-3-030-44534-8_26, jinguji2018fpga}. 


\subsubsection{\textbf{Key Generator Architecture}}
\begin{figure} [] 
	\centering
		\includegraphics[scale=.63] {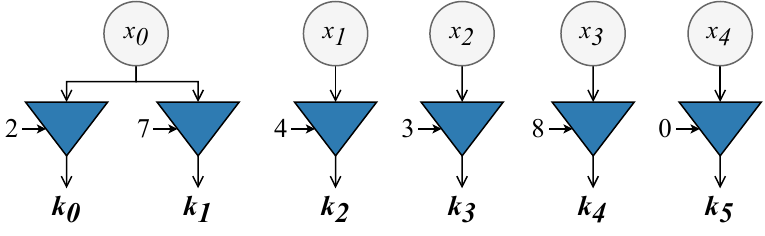} 
      \vspace{-4pt}
	\caption{The key generator layer for the example GBDT model in Fig.~\ref{fig:xgb}. The triangles are comparators.}
    \vspace{-8pt}
 \label{fig:key}
\end{figure}

The key generator is the first layer in the architecture. It generates a set of unique keys, which are used by decision nodes in the decision trees later. In our software tool, we traverse through all the decision trees and extract the features and their corresponding thresholds of the decision nodes within the whole GBDT model. In hardware, these comparisons are performed in parallel in the key generator layer using fully unrolled comparators. As a result, the inputs of this layer are a set of input features, each of which are $w\_feature$ bits. The outputs are a bundle of wires corresponding to the results of the comparators, each of which is one bit and denoted as $k_i$. It should be noted that only a set of unique keys is generated, and multiple instances of an input feature with the same threshold are generated as a single key. Fig.~\ref{fig:key} shows the corresponding key generator layer for the example GBDT model in Fig.~\ref{fig:xgb}. Note that in some instances, an input feature might be used in multiple comparisons (e.g., $x_0$ in Fig.~\ref{fig:key}), and in some instances, there could be features that are not involved in any comparisons.

\subsubsection{\textbf{Decision Tree Architecture}}
\begin{figure} [] 
	\centering
		\includegraphics[scale=.66] {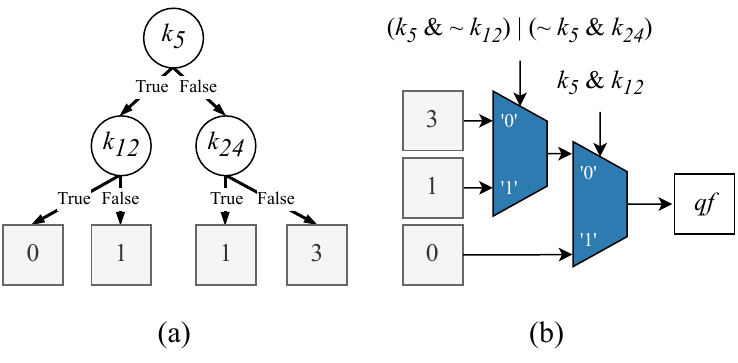} 
   \vspace{-4pt}
	\caption{The architecture of a decision tree.}
\vspace{-8pt}
 \label{fig:dt}
\end{figure}

Quantized decision trees are the second layer of the architecture. In a binary classification task, there are $M$ decision trees in total, whereas there are $N\times M$ decision trees in a multiclass classification task, where $N$ is the number of classes and $M$ is the number of decision trees per class. In our architecture, all decision trees are implemented fully unrolled and evaluated in parallel. The structure of a single quantized decision tree is depicted in Fig.~\ref{fig:dt} (a). As shown in the figure, a decision node uses $k_i$ as its key, which is generated in the previous layer. Furthermore, the leaf values are quantized into $w\_tree$ bits. The decision trees are implemented using a cascade of multiplexers that route the $qf$ value from a set of unique quantized leaf values. The select lines of the multiplexer are fed by the boolean expressions derived from each decision tree's structure. Fig.~\ref{fig:dt} (b) shows the hardware architecture of the decision tree of Fig.~\ref{fig:dt} (a). It can be seen that there are three unique leaf values in this decision tree: 0, 1, and 3. In our software tool, the combination of all paths that lead to each of these unique values is expressed as a single boolean function to feed the select line of the corresponding multiplexer. For example, two paths lead to the leaf value 1, and the corresponding boolean expression is $(k_5\&\sim k_{12}) | (\sim k_5\&k_{24})$. 

Encoding the paths and using multiplexers for decision trees is not a novel idea, as it is a straightforward way of implementing them. As discussed in Sec.~\ref{sec:pipelining}, our modular architecture and pipelining strategy allow FPGA design flows to optimize decision trees as BDD-like structures efficiently. In addition, as discussed in Sec.~\ref{sec:leaf}, our quantization scheme ensures that the minimum leaf value of each decision tree is 0 without any offsets, and the maximum leaf value of many decision trees fit into fewer number of bits than the target bitwidth $w\_feature$. These considerations during the quantization reduce the bitwidths of many decision trees and compress them further, resulting in lower hardware costs.

\subsubsection{\textbf{Adder Architecture}}
Adders are the last layer of the architecture. In a binary classification task, an adder module adds the leaf values returned by all decision trees plus the bias $qb$. However, since the final accumulation value is going to be compared against a threshold of zero to determine the binary label (negative accumulated value would mean Class 0, and positive accumulated value would mean Class 1), we can avoid adding the bias $qb$ in the adder architecture, and instead use it on the other side of the inequality as the classification threshold in the binary classification. In a multiclass classification task, however, there are $N$ separate adder modules that add the leaf values returned by the decision trees of each class in parallel, plus their corresponding bias $qb_n$. We use an adder tree structure within each adder module due to its efficiency and simplicity for implementing and pipelining.



\subsection{Pipelining}
\label{sec:pipelining}
The architecture of TreeLUT already exhibits a small delay, as it uses a modular architecture with a few layers. However, pipelining can reduce the critical path delay and increase the maximum clock frequency. 
Pipelining stages are inserted in our architecture such that an ideal initial interval (II) of 1 clock cycle is achieved. For this purpose, we consider three different locations for the registers, which can be controlled by the pipelining hyperparameters $[p_0, p_1, p_2]$. Here, $p_0$ and $p_1$ show if there are any pipelining stages after the key generator and decision tree modules, respectively. In addition, $p_2$ shows the number of pipelining stages in each adder tree module. As it implies, we do not put registers after each level in the adder trees, but the registers are evenly placed in a few levels based on the value of $p_2$. For example, if the depth of an adder tree is 6, and the value of $p_2$ is 1, only a single pipelining stage is inserted after the adders at the 3rd level. It is worth mentioning that we do not put any registers within the decision trees, as it would hinder the FPGA design flow from efficiently mapping them into LUTs. 

Our experiments indicate that the critical path delay is frequently due to adder trees needing to sum many operands. As a result, putting a single pipelining stage after the decision trees ($p_1 = 1$) or within the adder trees ($p_2 = 1$) usually results in a significant reduction in the critical path delay.

\section{Tool Flow}
\label{sec:toolflow}

\begin{figure} [] 
	\centering
		 \hspace{-5pt}\includegraphics[scale=.75]{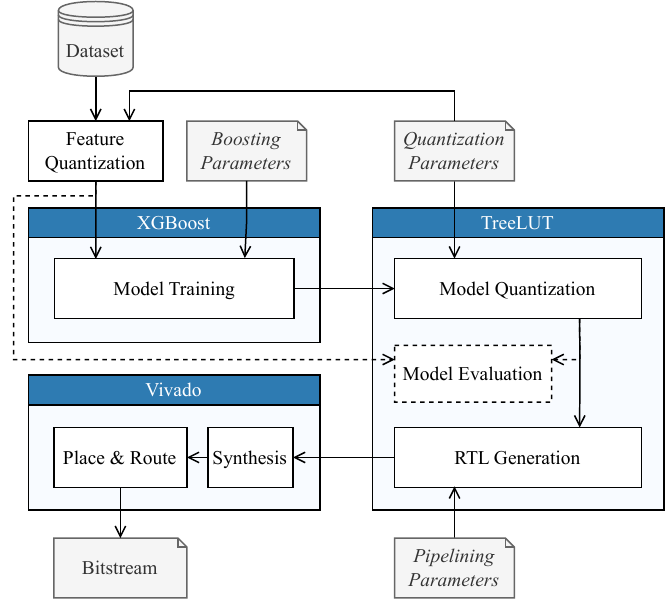} 
	\caption{The overview of the TreeLUT tool flow.}

 \label{fig:toolflow}
\end{figure}

We developed the TreeLUT tool as a Python library that can directly convert an XGBoost classifier model into RTL files in Verilog, given the quantization and pipelining parameters. The TreeLUT tool flow is shown in Fig.~\ref{fig:toolflow}.

As seen in the figure, the input features of the data are quantized (into $w\_feature$ bits), and the quantized dataset is fed into XGBoost for training a GBDT model, given the boosting parameters. Next, the leaves of the trained model are quantized (into $w\_tree$ bits) based on the TreeLUT quantization scheme. Finally, the quantized model and pipelining parameters are used to generate RTL files based on the TreeLUT architecture. In the TreeLUT Python library, we efficiently implemented the prediction function that can make predictions on validation and test sets given the quantization error introduced by the quantization parameters, and it models the exact behavior of hardware implementations in terms of accuracy.

Quantization parameters can be obtained in the same way as boosting parameters are found during training. For example, one may perform a grid or random search to tune boosting and quantization parameters simultaneously and evaluate them through our prediction function using a k-fold cross-validation algorithm or other evaluation techniques. Once the boosting and quantization parameters are tuned to satisfy the target accuracy, pipelining parameters can be set to maximize throughput and clock frequency. These parameters simply specify the boundaries of pipelining stages in the architecture based on the TreeLUT pipelining strategy. As we consider pipelining stages in a few positions in the architecture, the pipelining parameters create a very small design space, and users can tune them manually to obtain an efficient design given their target throughput and latency.

\section{Evaluation}
\label{sec:evaluation}

\begin{table*}[]
\caption{The boosting, quantization, and pipelining parameters used for the TreeLUT implementations.}
\vspace{-5pt}
\renewcommand{\arraystretch}{1.5}
\setlength{\tabcolsep}{0.2cm}
\resizebox{\linewidth}{!}{ 
\begin{tabular}{|c|l|c|cccc|cc|c|}
\hline
 & &  & \multicolumn{4}{c|}{\textbf{Boosting Parameters}} & \multicolumn{2}{c|}{\textbf{Quantization Parameters}}  & \\ \cline{4-9} 
\multirow{-2}{*}{\textbf{Dataset}} & \multicolumn{1}{c|}{\multirow{-2}{*}{\textbf{Method}}} & \multirow{-2}{*}{\textbf{Accuracy}} & \multicolumn{1}{c|}{\textbf{\textit{n\_estimators}}} & \multicolumn{1}{c|}{\textbf{\textit{max\_depth}}} & \multicolumn{1}{c|}{\textbf{\textit{eta}}} & \textbf{\textit{scale\_pos\_weight}} & \multicolumn{1}{c|}{\textbf{\textit{w\_feature}}} & \textbf{\textit{w\_tree}} & \multirow{-2}{*}{\textbf{Pipelining Parameters}} \\ \hline \hline
 & \cellcolor[HTML]{EFEFEF}TreeLUT (I) & \cellcolor[HTML]{EFEFEF}96.6\%  &  \cellcolor[HTML]{EFEFEF}30 & \cellcolor[HTML]{EFEFEF}5 & \cellcolor[HTML]{EFEFEF}0.8 & \cellcolor[HTML]{EFEFEF}\textemdash & \cellcolor[HTML]{EFEFEF}4 & \cellcolor[HTML]{EFEFEF}3  & \cellcolor[HTML]{EFEFEF}[0, 1, 1] \\ 
\multirow{-2}{*}{MNIST} & TreeLUT (II) & 95.6\% & 30 & 4 & 0.8 & \textemdash & 4 & 3  & [0, 1, 1] \\ \hline \hline
 & \cellcolor[HTML]{EFEFEF}TreeLUT (I) & \cellcolor[HTML]{EFEFEF}75.6\% & \cellcolor[HTML]{EFEFEF}13 & \cellcolor[HTML]{EFEFEF}5 & \cellcolor[HTML]{EFEFEF}0.8 & \cellcolor[HTML]{EFEFEF}\textemdash & \cellcolor[HTML]{EFEFEF}8 & \cellcolor[HTML]{EFEFEF}4  & \cellcolor[HTML]{EFEFEF}[0, 1, 1] \\
\multirow{-2}{*}{JSC} & TreeLUT (II) & 74.6\% & 10 & 5 & 0.3 & \textemdash & 8 & 2  & [0, 1, 0] \\  \hline \hline
 & \cellcolor[HTML]{EFEFEF}TreeLUT (I) & \cellcolor[HTML]{EFEFEF}92.7\% & \cellcolor[HTML]{EFEFEF}40 & \cellcolor[HTML]{EFEFEF}3 & \cellcolor[HTML]{EFEFEF}0.6 & \cellcolor[HTML]{EFEFEF}0.3 & \cellcolor[HTML]{EFEFEF}1 & \cellcolor[HTML]{EFEFEF}5  & \cellcolor[HTML]{EFEFEF}[0, 0, 1] \\
\multirow{-2}{*}{NID} & TreeLUT (II) & 91.5\% & 10 & 3 & 0.8 & 0.2 & 1 & 5  & [0, 0, 1] \\ \hline
\end{tabular}
}
\label{tbl:parameters}
\end{table*}

\begin{table}[]
\vspace{2pt}
\caption{Accuracy of pre-quantization floating-point GBDTs and post-quantization TreeLUT GBDTs.}
\vspace{-5pt}
\renewcommand{\arraystretch}{1.5}
\resizebox{\linewidth}{!}{ 
\begin{tabular}{|c|l|cc|}
\hline
 & \multicolumn{1}{c|}{} & \multicolumn{2}{c|}{\textbf{Accuracy}} \\ \cline{3-4} 
\multirow{-2}{*}{\textbf{Dataset}} & \multicolumn{1}{c|}{\multirow{-2}{*}{\textbf{Method}}} & \multicolumn{1}{c|}{\textbf{Before Quantization}} & \multicolumn{1}{c|}{\textbf{After Quantization}} \\ \hline \hline
 & \cellcolor[HTML]{EFEFEF}TreeLUT (I) & \cellcolor[HTML]{EFEFEF}96.9\% & \cellcolor[HTML]{EFEFEF}96.6\% \\
\multirow{-2}{*}{MNIST} & TreeLUT (II) & 96.5\% & 95.6\% \\ \hline \hline
 & \cellcolor[HTML]{EFEFEF}TreeLUT (I) & \cellcolor[HTML]{EFEFEF}75.7\% & \cellcolor[HTML]{EFEFEF}75.6\% \\
\multirow{-2}{*}{JSC} & TreeLUT (II) & 74.8\% & 74.6\% \\ \hline \hline
 & \cellcolor[HTML]{EFEFEF}TreeLUT (I) & \cellcolor[HTML]{EFEFEF}92.0\% & \cellcolor[HTML]{EFEFEF}92.7\% \\
\multirow{-2}{*}{NID} & TreeLUT (II) & 91.7\% & 91.5\% \\ \hline
\end{tabular}
}
\label{tbl:fpgbdts}
\end{table}

\begin{table}[]
\caption{Specifications of the classification datasets.}
\vspace{-5pt}
\renewcommand{\arraystretch}{1.5}
\setlength{\tabcolsep}{0.3cm}
\resizebox{\linewidth}{!}{ 
\begin{tabular}{|c| S[table-format=3.0]  S[table-format=2.0] |}
\hline
\multicolumn{1}{|l|}{\textbf{Dataset}} & \multicolumn{1}{l}{\textbf{Number of Input Features}} & \multicolumn{1}{l|}{\textbf{Number of Classes}} \\ \hline \hline
\cellcolor[HTML]{EFEFEF}MNIST & \cellcolor[HTML]{EFEFEF}784 & \cellcolor[HTML]{EFEFEF}10 \\ 
JSC & 16 & 5 \\ 
\cellcolor[HTML]{EFEFEF}NID & \cellcolor[HTML]{EFEFEF}593 & \cellcolor[HTML]{EFEFEF}2 \\ \hline
\end{tabular}
}
\label{tbl:dataset}

\end{table}

\begin{table*}[]
\caption{Comparison of TreeLUT with previous works in terms of accuracy and hardware costs. The accuracy results of the TreeLUT implementations were obtained from post-implementation functional simulations, and their hardware costs were obtained from post-place\&route utilization and timing reports. For the previous works, the results were quoted directly from their original papers.}
\vspace{-5pt}
\renewcommand{\arraystretch}{1.51}
\setlength{\tabcolsep}{0.2cm}
\resizebox{\linewidth}{!}{ 
\begin{tabular}{|c|l|c|c|S[table-format=6.0] S[table-format=6.0] S[table-format=2.0] S[table-format=3.0] S[table-format=3.0] c S[table-format=1.2e1, retain-zero-exponent = true] S[table-format=1.2e1, retain-zero-exponent = true] |}

\hline
 & \multicolumn{1}{c|}{} & &  & \multicolumn{8}{c|}{\textbf{Hardware Costs}} \\ \cline{5-12} 
\multirow{-2}{*}{\textbf{Dataset}} & \multicolumn{1}{c|}{\multirow{-2}{*}{\textbf{Method}}} & \multicolumn{1}{c|}{\multirow{-2}{*}{\textbf{Model}}} & \multirow{-2}{*} {\textbf{Accuracy}} & \multicolumn{1}{c|}{\textbf{LUT}} & \multicolumn{1}{c|}{\textbf{FF}} & \multicolumn{1}{c|}{\textbf{DSP}} & \multicolumn{1}{c|}{\textbf{BRAM}} & \multicolumn{1}{c|}{\textbf{F\textsubscript{max} (MHz)}} & \multicolumn{1}{c|}{\textbf{Latency (ns)}} & \multicolumn{1}{c|}{\textbf{Area~\texttimes~Delay}} & \textbf{\begin{tabular}[c]{@{}c@{}}Area~\texttimes~Delay\vspace{-6pt}\\ Ratio\vspace{1.5pt}\end{tabular}} \\ \hline \hline
  & \cellcolor[HTML]{EFEFEF}\textbf{TreeLUT (I)} &\bfseries \cellcolor[HTML]{EFEFEF}DT & \cellcolor[HTML]{EFEFEF}\bfseries 97\% & \cellcolor[HTML]{EFEFEF}\bfseries 4478 & \cellcolor[HTML]{EFEFEF}\bfseries 597 & \cellcolor[HTML]{EFEFEF}\bfseries 0 & \cellcolor[HTML]{EFEFEF}\bfseries 0 & \cellcolor[HTML]{EFEFEF}\bfseries 791 & \cellcolor[HTML]{EFEFEF}\bfseries 2.5 & \cellcolor[HTML]{EFEFEF}\bfseries 1.12e+4 & \cellcolor[HTML]{EFEFEF}\bfseries 1.39e+0 \\
 & {POLYBiNN (I)}~\cite{8596871} & DT & {97\%} & 109653 & \multicolumn{1}{r}{\textemdash} & \hspace*{0.18em}\textemdash & 0 & 100 & 90 & 9.87e+6 &  1.23e+3 \\
 & \cellcolor[HTML]{EFEFEF}\textbf{TreeLUT (II)} &\bfseries \cellcolor[HTML]{EFEFEF}DT & \cellcolor[HTML]{EFEFEF}\bfseries 96\% & \cellcolor[HTML]{EFEFEF}\bfseries 3499 & \cellcolor[HTML]{EFEFEF}\bfseries 759 & \cellcolor[HTML]{EFEFEF}\bfseries 0 & \cellcolor[HTML]{EFEFEF}\bfseries 0 & \cellcolor[HTML]{EFEFEF}\bfseries 874 & \cellcolor[HTML]{EFEFEF}\bfseries 2.3 & \cellcolor[HTML]{EFEFEF}\bfseries 8.05e+3 & \cellcolor[HTML]{EFEFEF}\bfseries 1.00e+0 \\
 & {POLYBiNN (II)}~\cite{8596871} & DT & {96\%} & 9943 & \multicolumn{1}{r}{\textemdash}  &  \hspace*{0.18em}\textemdash& 0 & 100 & 70 & 6.96e+5 &  8.65e+1 \\
 & \cellcolor[HTML]{EFEFEF}{PolyLUT-Add}~\cite{10705569} & \cellcolor[HTML]{EFEFEF}NN & \cellcolor[HTML]{EFEFEF}{96\%} & \cellcolor[HTML]{EFEFEF}14810 & \cellcolor[HTML]{EFEFEF}2609 & \cellcolor[HTML]{EFEFEF}0 & \cellcolor[HTML]{EFEFEF}0 & \cellcolor[HTML]{EFEFEF}625 & \cellcolor[HTML]{EFEFEF}10 & \cellcolor[HTML]{EFEFEF}1.48e+5 & \cellcolor[HTML]{EFEFEF} 1.84e+1 \\
 & {NeuraLUT}~\cite{10705559} & NN & {96\%} & 54798 & 3757 & 0 & 0 & 431 & 12 & 6.58e+5 &  8.17e+1 \\
 & \cellcolor[HTML]{EFEFEF}{PolyLUT}~\cite{10416099} & \cellcolor[HTML]{EFEFEF}NN & \cellcolor[HTML]{EFEFEF}{96\%} & \cellcolor[HTML]{EFEFEF}70673 & \cellcolor[HTML]{EFEFEF}4681 & \cellcolor[HTML]{EFEFEF}0 & \cellcolor[HTML]{EFEFEF}0 & \cellcolor[HTML]{EFEFEF}378 & \cellcolor[HTML]{EFEFEF}16 & \cellcolor[HTML]{EFEFEF}1.13e+6 & \cellcolor[HTML]{EFEFEF} 1.41e+2 \\
 & {FINN}~\cite{10.1145/3020078.3021744} & NN & {96\%} & 91131 & \multicolumn{1}{r}{\textemdash} & 0 & 5 & 200 & 310 & 2.83e+7 &  3.51e+3 \\
\multirow{-9}{*}{{MNIST}} & \cellcolor[HTML]{EFEFEF}{hls4ml (Ngadiuba et al.)}~\cite{Ngadiuba_2021} & \cellcolor[HTML]{EFEFEF}NN & \cellcolor[HTML]{EFEFEF}{95\%} & \cellcolor[HTML]{EFEFEF}260092 & \cellcolor[HTML]{EFEFEF}165513 & \cellcolor[HTML]{EFEFEF}0 & \cellcolor[HTML]{EFEFEF}345 & \cellcolor[HTML]{EFEFEF}200 & \cellcolor[HTML]{EFEFEF}190 & \cellcolor[HTML]{EFEFEF}4.94e+7 & \cellcolor[HTML]{EFEFEF} 6.14e+3 \\ \hline \hline
 & \textbf{TreeLUT (I)} &\bfseries DT & \bfseries 76\% & \bfseries 2234 & \bfseries 347 & \bfseries 0 & \bfseries 0 & \bfseries 735 & \bfseries 2.7 & \bfseries 6.03e+3 &  \bfseries 6.89e+0 \\
 & \cellcolor[HTML]{EFEFEF}{hls4ml (Fahim et al.)}~\cite{fahim2021hls4ml} & \cellcolor[HTML]{EFEFEF}NN & \cellcolor[HTML]{EFEFEF}{76\%} & \cellcolor[HTML]{EFEFEF}63251 & \cellcolor[HTML]{EFEFEF}4394 & \cellcolor[HTML]{EFEFEF}38 & \cellcolor[HTML]{EFEFEF}0 & \cellcolor[HTML]{EFEFEF}200 & \cellcolor[HTML]{EFEFEF}45 & \cellcolor[HTML]{EFEFEF}2.85e+6 & \cellcolor[HTML]{EFEFEF} 3.25e+3 \\
 & \textbf{TreeLUT (II)} &\bfseries DT &\bfseries 75\% &\bfseries 796 &\bfseries 74 &\bfseries 0 &\bfseries 0 &\bfseries 887 &\bfseries 1.1 &\bfseries 8.76e+2 &\bfseries  1.00e+0 \\
 & \cellcolor[HTML]{EFEFEF}{Alsharari et al.}~\cite{10652593} & \cellcolor[HTML]{EFEFEF}DT & \cellcolor[HTML]{EFEFEF}{75\%} & \cellcolor[HTML]{EFEFEF}6500 & \multicolumn{1}{r}{\cellcolor[HTML]{EFEFEF} \textemdash} & \cellcolor[HTML]{EFEFEF}0 & \cellcolor[HTML]{EFEFEF}0 & \cellcolor[HTML]{EFEFEF}670 & \cellcolor[HTML]{EFEFEF}7.1 & \cellcolor[HTML]{EFEFEF}4.62e+4 & \cellcolor[HTML]{EFEFEF} 5.27e+1 \\
 & {PolyLUT-Add}~\cite{10705569} & NN & {75\%} & 36484 & 1209 & 0 & 0 & 315 & 16 & 5.84e+5 &  6.67e+2 \\
 & \cellcolor[HTML]{EFEFEF}{NeuraLUT}~\cite{10705559} & \cellcolor[HTML]{EFEFEF}NN & \cellcolor[HTML]{EFEFEF}{75\%} & \cellcolor[HTML]{EFEFEF}92357 & \cellcolor[HTML]{EFEFEF}4885 & \cellcolor[HTML]{EFEFEF}0 & \cellcolor[HTML]{EFEFEF}0 & \cellcolor[HTML]{EFEFEF}368 & \cellcolor[HTML]{EFEFEF}14 & \cellcolor[HTML]{EFEFEF}1.29e+6 & \cellcolor[HTML]{EFEFEF} 1.48e+3 \\
 & {PolyLUT}~\cite{10416099} & NN & {75\%} & 236541 & 2775 & 0 & 0 & 235 & 21 & 4.97e+6 &  5.67e+3 \\
 & \cellcolor[HTML]{EFEFEF}{hls4ml (Summers et al.)}~\cite{Summers_2020} & \cellcolor[HTML]{EFEFEF}DT & \cellcolor[HTML]{EFEFEF}{74\%} & \cellcolor[HTML]{EFEFEF}96148 & \cellcolor[HTML]{EFEFEF}42802 & \cellcolor[HTML]{EFEFEF}0 & \cellcolor[HTML]{EFEFEF}0 & \cellcolor[HTML]{EFEFEF}200 & \cellcolor[HTML]{EFEFEF}60 & \cellcolor[HTML]{EFEFEF}5.77e+6 & \cellcolor[HTML]{EFEFEF} 6.59e+3 \\
\multirow{-9}{*}{{JSC}} & {LogicNets}~\cite{9221584} & NN & {72\%} & 37900 & \multicolumn{1}{r}{\textemdash} & 0 & 0 & 384 & 13 & 4.93e+5 &  5.63e+2 \\ \hline \hline
 & \cellcolor[HTML]{EFEFEF}\textbf{TreeLUT (I)} &\bfseries \cellcolor[HTML]{EFEFEF}DT & \cellcolor[HTML]{EFEFEF}\bfseries 93\% & \cellcolor[HTML]{EFEFEF}\bfseries 345 & \cellcolor[HTML]{EFEFEF}\bfseries 33 & \cellcolor[HTML]{EFEFEF}\bfseries 0 & \cellcolor[HTML]{EFEFEF}\bfseries 0 & \cellcolor[HTML]{EFEFEF}\bfseries 681 & \cellcolor[HTML]{EFEFEF}\bfseries 1.5 & \cellcolor[HTML]{EFEFEF}\bfseries 5.18e+2 & \cellcolor[HTML]{EFEFEF}\bfseries 5.81e+0 \\
 & \textbf{TreeLUT (II)} &\bfseries DT &\bfseries 92\% &\bfseries 89 &\bfseries 19 &\bfseries 0 &\bfseries 0 &\bfseries 1047 &\bfseries 1.0 &\bfseries 8.90e+1 &\bfseries  1.00e+0 \\
 & \cellcolor[HTML]{EFEFEF}{Alsharari et al. (I)}~\cite{10652593} & \cellcolor[HTML]{EFEFEF}DT & \cellcolor[HTML]{EFEFEF}{92\%} & \cellcolor[HTML]{EFEFEF}1800 & \multicolumn{1}{r}{\cellcolor[HTML]{EFEFEF} \textemdash} & \cellcolor[HTML]{EFEFEF}0 & \cellcolor[HTML]{EFEFEF}0 & \cellcolor[HTML]{EFEFEF}714 & \cellcolor[HTML]{EFEFEF}6.9 & \cellcolor[HTML]{EFEFEF}1.24e+4 & \cellcolor[HTML]{EFEFEF} 1.40e+2 \\
 & {Alsharari et al. (II)}~\cite{10652593} & DT & {92\%} & 170 & \multicolumn{1}{r}{\textemdash} & 0 & 0 & 724 & 1.4 & 2.38e+2 &  2.67e+0 \\
 & \cellcolor[HTML]{EFEFEF}{PolyLUT-Add}~\cite{10705569} & \cellcolor[HTML]{EFEFEF}NN & \cellcolor[HTML]{EFEFEF}{92\%} & \cellcolor[HTML]{EFEFEF}1649 & \cellcolor[HTML]{EFEFEF}830 & \cellcolor[HTML]{EFEFEF}0 & \cellcolor[HTML]{EFEFEF}0 & \cellcolor[HTML]{EFEFEF}620 & \cellcolor[HTML]{EFEFEF}8 & \cellcolor[HTML]{EFEFEF}1.32e+4 & \cellcolor[HTML]{EFEFEF} 1.48e+2 \\
 & {PolyLUT}~\cite{10416099} & NN & {92\%} & 3336 & 686 & 0 & 0 & 529 & 9 & 3.00e+4 &  3.37e+2 \\
 & \cellcolor[HTML]{EFEFEF}{Murovic et al.}~\cite{MUROVIC20211} & \cellcolor[HTML]{EFEFEF}NN & \cellcolor[HTML]{EFEFEF}{92\%} & \cellcolor[HTML]{EFEFEF}17990 & \cellcolor[HTML]{EFEFEF}0 & \cellcolor[HTML]{EFEFEF}0 & \cellcolor[HTML]{EFEFEF}0 & \cellcolor[HTML]{EFEFEF}55 & \cellcolor[HTML]{EFEFEF}18 & \cellcolor[HTML]{EFEFEF}3.24e+5 & \cellcolor[HTML]{EFEFEF} 3.64e+3 \\
\multirow{-8}{*}{{NID}} & {LogicNets}~\cite{9221584} & NN & {91\%} & 15900 & \multicolumn{1}{r}{\textemdash} & 0 & 0 & 471 & 11 & 1.75e+5 &  1.97e+3 \\ \hline
\end{tabular}

}
\label{tbl:main}

\end{table*}

We evaluated the efficiency of our TreeLUT method against previous works on the implementations of neural networks and decision trees on FPGAs, including DWN~\cite{pmlr-v235-bacellar24a}, POLYBiNN~\cite{8596871}, PolyLUT-Add~\cite{10705569}, NeuraLUT~\cite{10705559}, PolyLUT~\cite{10416099}, FINN~\cite{10.1145/3020078.3021744}, hls4ml~\cite{Ngadiuba_2021, fahim2021hls4ml, Summers_2020}, Alsharari et al.~\cite{10652593}, Murovic et al.~\cite{MUROVIC20211}, and LogicNets~\cite{9221584}. As in previous works, we evaluated TreeLUT on the MNIST, JSC, and NID datasets, which are commonly used to evaluate ultra-low latency and area architectures~\cite{10416099}. We used the same JSC dataset as used by~\cite{10652593, fahim2021hls4ml, Summers_2020} and the same NID dataset as used by~\cite{10705569, 10416099,9221584}. The specifications of each dataset are provided in Table~\ref{tbl:dataset}. 

In what follows, we discuss the training, quantization, and hardware implementation of the TreeLUT models in detail and provide the evaluation results. Finally, we discuss the results and compare our method to the mentioned previous works.

\subsection{Model Training and Quantization}
As discussed in Section~\ref{sec:toolflow}, the TreeLUT tool is built on top of XGBoost~\cite{10.1145/2939672.2939785} as a scalable machine learning library for GBDTs. Although it supports a wide range of hyperparameters, we considered tuning the following ones for the evaluation of TreeLUT.

\begin{itemize}[left=0pt]
\item{\textbf{\textit{n\_estimators}:}}  The number of decision trees. It uses \textit{n\_estimators} decision trees per class in multiclass classification but uses\linebreak\textit{n\_estimators} decision trees in total in binary classification.
\item{\textbf{\textit{max\_depth}:}} The maximum depth of each decision tree.
\item{\textbf{\textit{eta}:}} Shrinking weights in update to prevent overfitting.
\item{\textbf{\textit{scale\_pos\_weight}:}} Controlling the balance of weights in binary classification with class imbalance.
\end{itemize}

Besides the aforementioned XGBoost boosting parameters, the TreeLUT quantization parameters \textbf{\textit{w\_feature}} and \textbf{\textit{w\_tree}} need to be tuned to ensure our target accuracy. We performed a grid search on a subset of values for each parameter to reach the desired target accuracy. As different parameters end up getting different accuracy levels and model complexities, we considered two target accuracy values for each dataset to show the scalability of our method. These two designs are labeled as TreeLUT (I) and TreeLUT (II). The boosting and quantization parameters for each case are summarized in Table~\ref{tbl:parameters}. In addition,  Table~\ref{tbl:fpgbdts} shows the accuracy of GBDTs before and after quantization of thresholds and leaves. These accuracy values are based on the chosen hyperparameter values, and higher accuracy may be achieved by tuning additional hyperparameters and expanding the search space while considering the trade-off between training time, accuracy, and hardware costs.

\begin{table*}[]
\caption{Comparison of TreeLUT with DWN~\cite{pmlr-v235-bacellar24a} in terms of accuracy and hardware costs. For DWN, the results were quoted directly from its original paper~\cite{pmlr-v235-bacellar24a}.}

\vspace{-5pt}
\renewcommand{\arraystretch}{1.4}
\setlength{\tabcolsep}{0.28cm}
\resizebox{\linewidth}{!}{ 
\begin{tabular}{|c|l|c|c|S[table-format=4.0] S[table-format=4.0] c c c c S[table-format=1.2e1, retain-zero-exponent = true] S[table-format=1.2e1, retain-zero-exponent = true] |}

\hline
 & \multicolumn{1}{c|}{} & &  & \multicolumn{8}{c|}{\textbf{Hardware Costs}} \\ \cline{5-12} 
\multirow{-2}{*}{\textbf{Dataset}} & \multicolumn{1}{c|}{\multirow{-2}{*}{\textbf{Method}}} & \multicolumn{1}{c|}{\multirow{-2}{*}{\textbf{Model}}} & \multirow{-2}{*} {\textbf{Accuracy}} & \multicolumn{1}{c|}{\textbf{LUT}} & \multicolumn{1}{c|}{\textbf{FF}} & \multicolumn{1}{c|}{\textbf{DSP}} & \multicolumn{1}{c|}{\textbf{BRAM}} & \multicolumn{1}{c|}{\textbf{F\textsubscript{max} (MHz)}} & \multicolumn{1}{c|}{\textbf{Latency (ns)}} & \multicolumn{1}{c|}{\textbf{Area~\texttimes~Delay}} & \textbf{\begin{tabular}[c]{@{}c@{}}Area~\texttimes~Delay\vspace{-6pt}\\ Ratio\vspace{1.5pt}\end{tabular}} \\ \hline \hline
\multicolumn{1}{|c|}{} & \multicolumn{1}{l|}{\cellcolor[HTML]{EFEFEF}DWN~\cite{pmlr-v235-bacellar24a}} & \multicolumn{1}{c|}{\cellcolor[HTML]{EFEFEF}NN} & \multicolumn{1}{c|}{\cellcolor[HTML]{EFEFEF}97.8\%} & \cellcolor[HTML]{EFEFEF}2092 & \cellcolor[HTML]{EFEFEF}1757 & \cellcolor[HTML]{EFEFEF}0 & \cellcolor[HTML]{EFEFEF}0 & \cellcolor[HTML]{EFEFEF}873 & \cellcolor[HTML]{EFEFEF}9.2 & \cellcolor[HTML]{EFEFEF}1.92e+4 & \multicolumn{1}{c|}{\cellcolor[HTML]{EFEFEF}4.02e+0} \\
\multicolumn{1}{|c|}{\multirow{-2}{*}{MNIST}} & \multicolumn{1}{l|}{\textbf{TreeLUT}} & \multicolumn{1}{c|}{\textbf{DT}} & \multicolumn{1}{c|}{\bfseries{96.6\%}} & \bfseries 2278 & \bfseries 597 & \bfseries 0 & \bfseries 0 & \bfseries 974 & \bfseries 2.1 & \bfseries 4.78e+3 & \bfseries 1.00e+0 \\ \hline
 \hline
\multicolumn{1}{|c|}{} & \multicolumn{1}{l|}{\cellcolor[HTML]{EFEFEF}\textbf{TreeLUT}} & \multicolumn{1}{c|}{\cellcolor[HTML]{EFEFEF}\textbf{DT}} & \multicolumn{1}{c|}{\cellcolor[HTML]{EFEFEF}\textbf{75.6\%}} & \cellcolor[HTML]{EFEFEF}\bfseries 1192 & \cellcolor[HTML]{EFEFEF}\bfseries 347 & \cellcolor[HTML]{EFEFEF}\bfseries 0 & \cellcolor[HTML]{EFEFEF}\bfseries 0 & \cellcolor[HTML]{EFEFEF}\bfseries 962 & \cellcolor[HTML]{EFEFEF}\bfseries 2.1 & \cellcolor[HTML]{EFEFEF}\bfseries 2.50e+3 & \multicolumn{1}{c|}{\cellcolor[HTML]{EFEFEF}\bfseries 1.00e+0} \\
\multicolumn{1}{|c|}{\multirow{-2}{*}{JSC}} & \multicolumn{1}{l|}{DWN~\cite{pmlr-v235-bacellar24a}} & \multicolumn{1}{c|}{NN} & \multicolumn{1}{c|}{75.6\%} & 2144 & 1457 & 0 & 0 & 903 & 8.9 & 1.91e+4 & 7.62e+0 \\ \hline
\end{tabular}
}
\label{tbl:dwn}

\end{table*}

\subsection{Hardware Implementation}

After training and quantizing the GBDT models using the XGBoost and TreeLUT libraries, our tool generates RTL files in Verilog. All designs were synthesized and placed \& routed using Vivado 2023.1. For consistency with previous works~\cite{10705569, 10705559, 10416099, 9221584}, we used the xcvu9p-flgb2104-2-i FPGA part with the Flow\_PerfOptimized\_high settings in the Out-of-Context (OOC) synthesis mode. It is worth noting that the TreeLUT tool took a few seconds to quantize a given XGBoost model, test it for accuracy, and convert it into RTL code. In comparison, some LUT-based NN tools, such as PolyLUT-Add and PolyLUT, might take hours to generate RTL code~\cite{10705569}.

As discussed in Section~\ref{sec:toolflow}, TreeLUT can pipeline designs given pipelining parameters with an II (initial interval) of 1 clock cycle. The pipelining parameters that we used for each design are summarized in Table~\ref{tbl:parameters}. The parameters are formatted as $[p0, p1, p2]$, as described in Sec.~\ref{sec:pipelining}.
In our implementations, the pipelining parameters were set to minimize the area-delay product. However, one could use more pipelining stages to increase the clock frequency and throughput further at the possible expense of higher latency. In Vivado, we set the target clock period to be in the range of 1 to 1.5 ns and input/output delay with the range of 0.05 to 0.1 ns, considering the complexity of each design.

Table~\ref{tbl:main} shows evaluation results of TreeLUT and previous works in terms of accuracy and hardware costs. In the table, Area $\times$ Delay is the multiplication of the LUT count and latency, and the last column shows the ratio of Area $\times$ Delay across all previous works for the given dataset. The column labeled ``Model" shows which methods are based on NNs (neural networks) or DTs (decision trees). Similarly, Fig.~\ref{fig:results} compares TreeLUT with the previous works listed in Table~\ref{tbl:main} in terms of area-delay product hardware cost and accuracy. The accuracy results of the TreeLUT implementations were obtained from post-implementation functional simulations, and the hardware costs were obtained from post-place\&route utilization and timing reports. In Table~\ref{tbl:main} and Fig.~\ref{fig:results}, the accuracy values were rounded to the nearest integers in percentages for consistency with some previous works. However, the more precise accuracy values of TreeLUT can be found in Table~\ref{tbl:parameters} and Table~\ref{tbl:fpgbdts}. For previous works, we quoted the results from their original papers.
If multiple implementations were present for a single dataset, we reported the results of the most comparable one. In other words, we chose those implementations that had closer accuracy values to our TreeLUT implementations to ensure a fair comparison of hardware costs. We similarly reported their post-place\&route implementation results if available. Otherwise, the post-synthesis results were reported.
\vspace{-2pt}

\subsection{Discussion}
\begin{figure}[]
    \centering
        \begin{subfigure}[t]{0.48\textwidth}
        \centering
        \includegraphics[width=\textwidth]{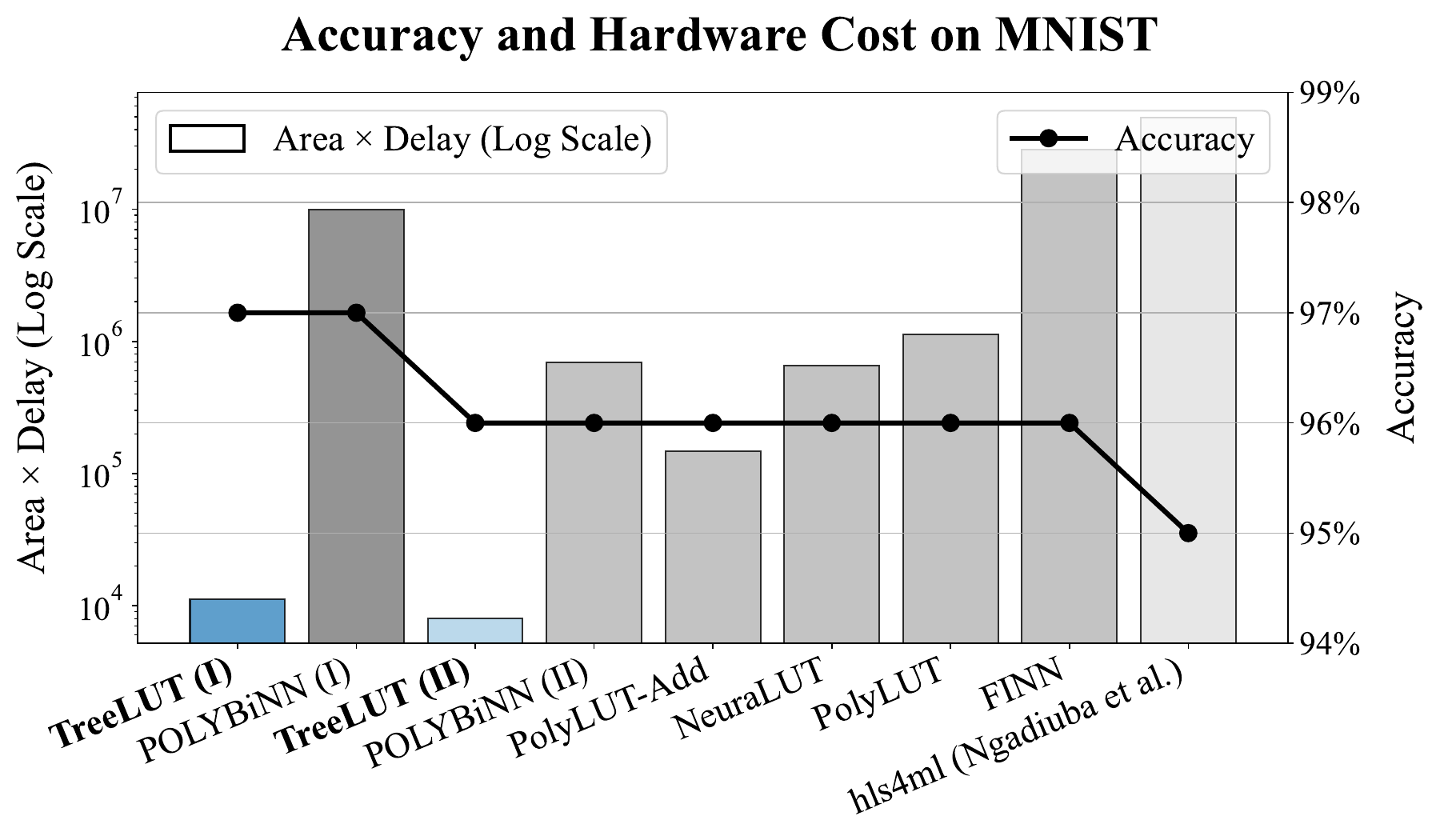} 

    \end{subfigure}
    \centering
        \begin{subfigure}[t]{0.48\textwidth}
        \centering
        \includegraphics[width=\textwidth]{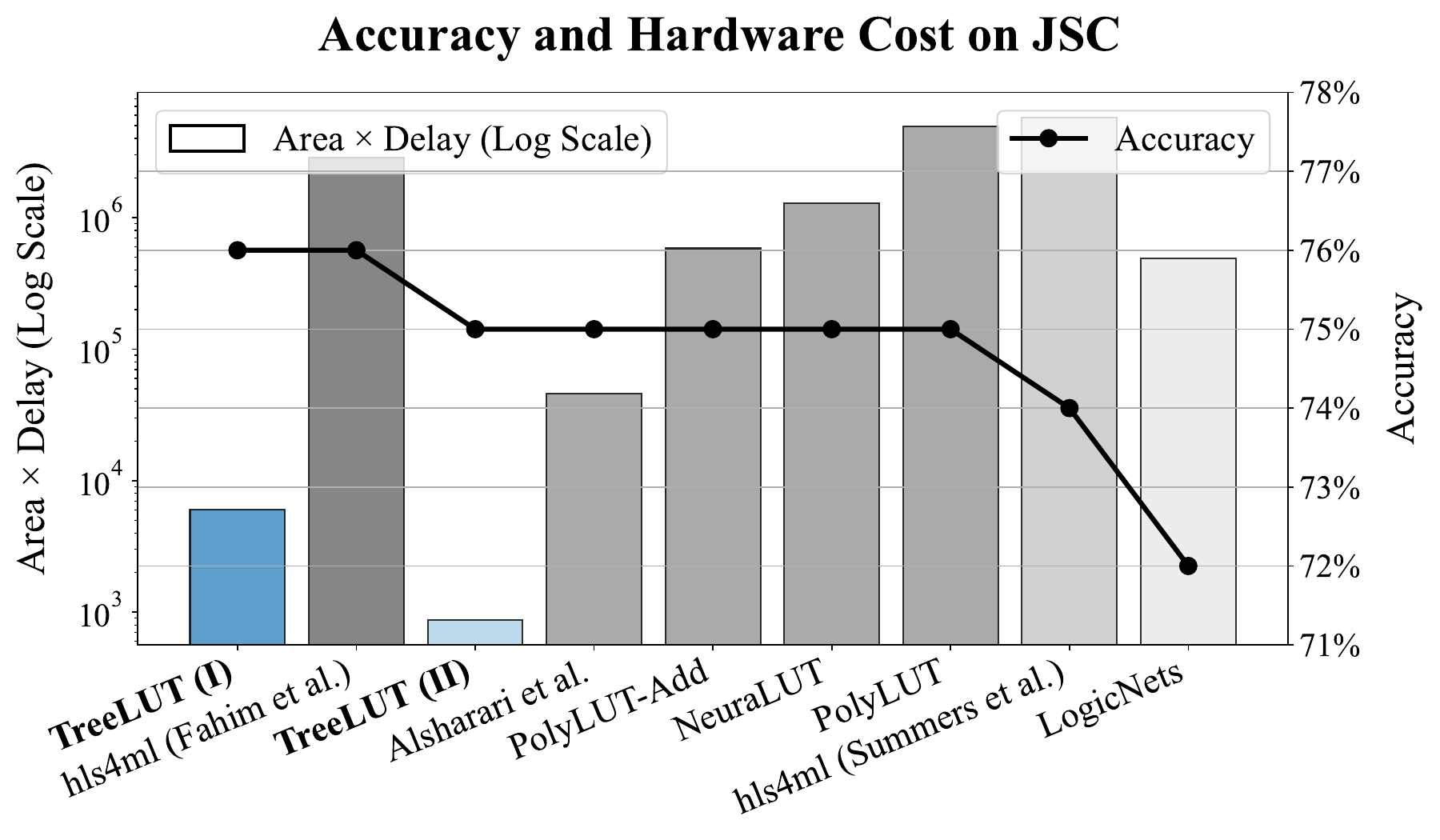}
    \end{subfigure}\hfill
    \begin{subfigure}[t]{0.48\textwidth}
        \centering
        \includegraphics[width=\textwidth]{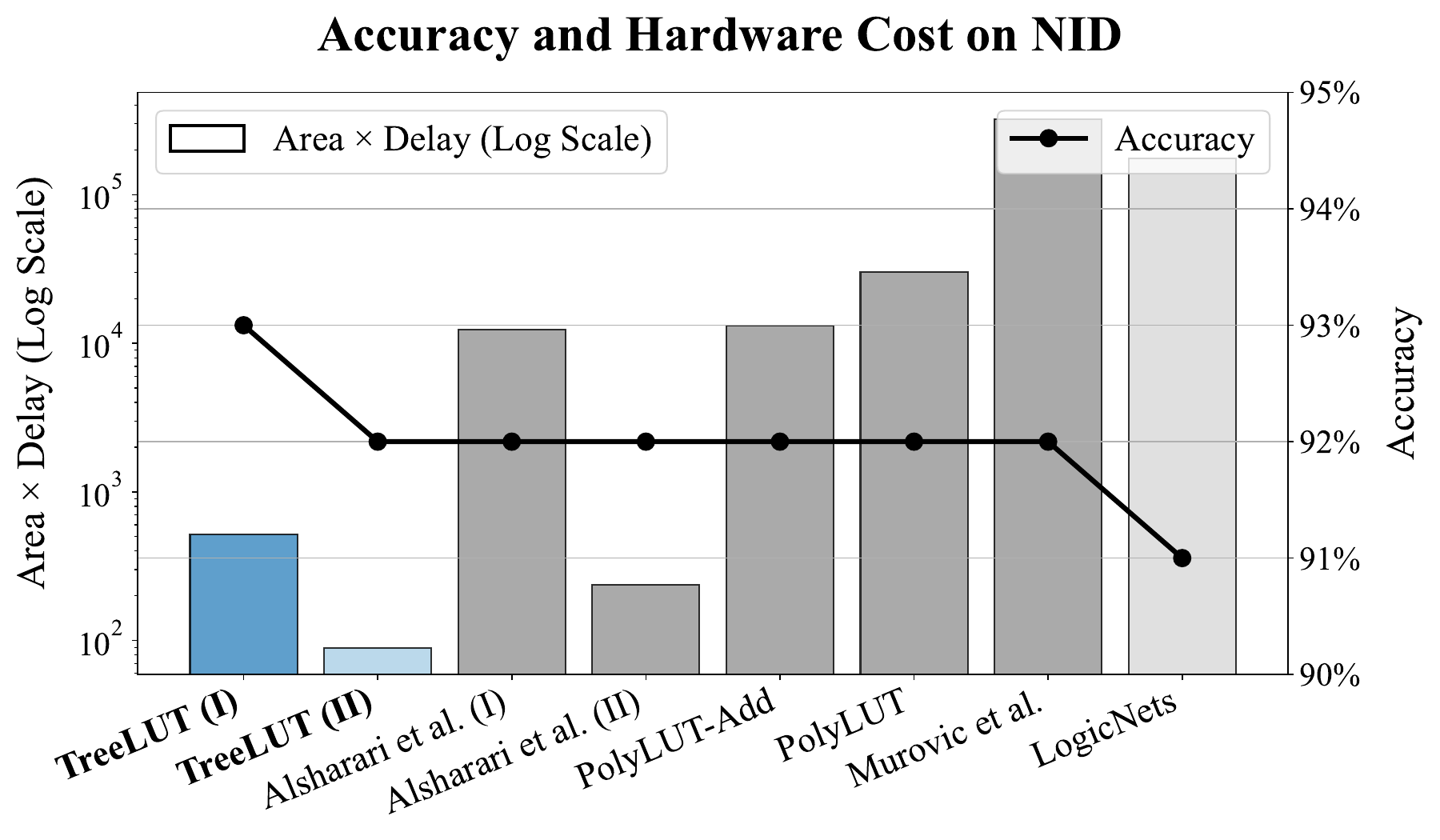}
    \end{subfigure}\hfill

    \caption{Comparison of TreeLUT with previous works. The bar charts show the area-delay product values using a logarithmic scale (the left y-axis), and the line charts show the test accuracy values (the right y-axis).}
   \vspace{-8pt}
    \label{fig:results}
\end{figure} 

As seen in Table~\ref{tbl:main} and Fig.~\ref{fig:results}, there are two TreeLUT implementations for each dataset, which are labeled as TreeLUT (I) and TreeLUT (II). The table shows that TreeLUT outperforms the listed previous works in terms of area, throughput, and latency at the same or higher accuracy on each dataset. The comparison of TreeLUT with DWN will be discussed later in a separate table. In what follows, we compare our implementation results against the state-of-the-art results on each dataset at comparable accuracy values.

On the MNIST dataset, TreeLUT (I) achieves an accuracy of around 97\%, similar to POLYBiNN (I) and higher than the other previous works in Table~\ref{tbl:main}. In comparison to POLYBiNN (I), TreeLUT (I) delivers 7.9$\times$ higher maximum frequency with 36$\times$ lower latency, utilizing 24.5$\times$ fewer LUTs, which results in a 881.5$\times$ reduction in area-delay product. 
To perform an apples-to-apples comparison to the rest of the previous works, we relaxed the accuracy target of TreeLUT to approximately match the accuracy of other works, resulting in the TreeLUT (II) architecture. PolyLUT-Add -- a recent LUT-based NN method -- has the lowest area-delay product among the previous methods. Compared to PolyLUT-Add, TreeLUT (II) delivers 1.4$\times$ higher maximum frequency with 4.3$\times$ lower latency, utilizing 4.2$\times$ fewer LUTs, which results in an 18.4$\times$ reduction in area-delay product.

On JSC, TreeLUT (I) achieves an accuracy of around 76\%, and TreeLUT (II) achieves an accuracy of around 75\%. At similar accuracy values, respectively, hls4ml (Fahim et al.) and Alsharari et al. have the lowest area-delay product among the previous works. In comparison with hls4ml (Fahim et al.), TreeLUT (I) delivers 3.7$\times$ higher maximum frequency with 16.7$\times$ lower latency, utilizing 28.3$\times$ fewer LUTs. In addition, hls4ml (Fahim et al.) utilizes 38 DSPs, whereas TreeLUT utilizes no DSPs. Setting aside DSP resources, TreeLUT (I) still results in  471.9$\times$ lower area-delay product. In comparison with Alsharari et al., TreeLUT (II) delivers 1.3$\times$ higher maximum frequency with 6.5$\times$ lower latency, utilizing 8.2$\times$ fewer LUTs, which results in a 52.7$\times$ reduction in area-delay product.

On the NID dataset, TreeLUT (I) achieves an accuracy of around 93\%, higher than all the previous works. Relaxing the accuracy target, TreeLUT (II) achieves an accuracy of around 92\%, which makes it comparable with the listed previous works. At a similar accuracy, Alsharari et al. (II) has the lowest area-delay product among all the previous methods. In comparison, TreeLUT (II) delivers 1.4$\times$ higher maximum frequency with 1.4$\times$ lower latency and utilizes 1.9$\times$ fewer LUTs, which results in a 2.7$\times$ reduction in area-delay product. In terms of scalability, Alsharari et al. (I) uses an 8-bit integer precision to achieve an accuracy of 92\%, whereas Alsharari et al. (II) uses a binary precision that results in an accuracy of 91.5\%. As seen, there is a significant gap between these two implementations in terms of area and latency, while the improvement in accuracy is 0.5\%. On the other hand, the accuracy improves from 91.5\% to 92.7\% when we switch from TreeLUT (I) to TreeLUT (II), while the hardware cost increase is less significant compared to Alsharari et al. In addition, TreeLUT quantizes thresholds and leaves using different bitwidths without using quantization-aware training techniques unlike Alsharari et al. Moreover, the TreeLUT tool directly generates RTL files without relying on external tools such as HLS, which makes it more scalable for complex tasks, as discussed in~\cite{Summers_2020}. Apart from the efficiency, simplicity, and scalability, it is worth noting that a factor that resulted in this significant difference in the hardware costs between TreeLUT and Alsharari et al. was that Alsharari et al. used more complex GBDT models with larger number of decision trees in their JSC and NID implementations, which was either due to the limitations of their quantization scheme that possibly required more decision trees to compensate for quantization error or due to insufficient tuning of the boosting parameters.

DWN~\cite{pmlr-v235-bacellar24a} is a recent LUT-based NN work. In this method, input data is binarized using a distributive thermometer encoding scheme~\cite{bacellar2022distributive}, consisting of a set of comparisons with thresholds. Although TreeLUT and DWN follow different processes during training and inference, TreeLUT performs a set of threshold comparisons on input data as well. DWN performs the threshold comparisons as a preprocessing step offline, whereas TreeLUT performs such operations directly on FPGAs using a key generator layer. For this reason and to make an apples-to-apples comparison with DWN, we bypassed the key generator layers of the TreeLUT (I) designs and provided the implementation results in Table~\ref{tbl:dwn}. As the threshold comparisons are assumed to be performed offline for this evaluation, we could have ignored the threshold quantization step of TreeLUT and trained GBDT models from scratch using floating-point data and tuned the hyperparameters accordingly to get higher accuracy values. However, we settled for using the TreeLUT (I) designs with the key generator layers bypassed as the only modification. On MNIST, DWN has 1.2\% higher accuracy than TreeLUT (I), resulting in a 4$\times$ increase in area-delay product. On JSC, however, DWN and TreeLUT (I) have the same accuracy, while TreeLUT (I) resulted in a 7.6$\times$ reduction in area-delay product. It is worth noting that the hardware architecture of DWN seems to have an argmax layer, which is in contrast to TreeLUT and other LUT-based NN works. However, the hardware resource utilization of the argmax layer is trivial compared to the overall architecture, as it can be implemented using a few simple comparators.

\section{Conclusion}
In this work, we proposed TreeLUT as an open-source tool for implementing GBDTs as an alternative to DNNs using a combination of efficient quantization scheme, hardware architecture, and pipelining strategy. Evaluated on multiple classification tasks, TreeLUT demonstrated lower area-delay product hardware cost at competitive accuracy compared to existing methods.

\begin{acks}
This work was supported in part by the National Science Foundation under Grant No. PFI-TT 2016390 and by the University of Minnesota's
Doctoral Dissertation Fellowship. We thank the anonymous shepherd and all the reviewers for their valuable feedback, which helped improve the quality of this work.
\end{acks}



\bibliographystyle{ACM-Reference-Format}
\balance
\bibliography{mainBib}

\appendix

\end{document}